\documentclass[letterpaper, 10 pt, conference]{ieeeconf}  

\IEEEoverridecommandlockouts                              

\usepackage{graphics} 
\usepackage{graphicx}
\usepackage{cite}
\usepackage{amsmath} 
\usepackage{amssymb}  
\usepackage{bm}      
\usepackage[normalem]{ulem}

\title{\LARGE \bf
Optimal Inverted Landing in a Small Aerial Robot
with Varied Approach Velocities and Landing Gear Designs }

\author{Bryan Habas$^{1}$, Bader AlAttar$^{1}$, Brian Davis$^{1}$, Jack W. Langelaan$^{2},$ \textit{IEEE, Member}, Bo Cheng$^{1}$, \textit{IEEE, Member}
\thanks{$^{1}$Biological and Robotic Intelligent Fluid Locomotion Lab, Department of Mechanical Engineering, The Pennsylvania State University, University Park, PA 16802, USA. Corresponding to B.C. {\tt\small buc10@psu.edu}}
\thanks{$^{2}$Air Vehicle Intelligence and Autonomy Lab, Department of Aerospace Engineering, The Pennsylvania State University, University Park, PA 16802, USA}
}

\begin{document}

\maketitle
\thispagestyle{empty}
\pagestyle{empty}
\hbadness=500000 
\vbadness=600000

\begin{abstract}

    Inverted landing is a challenging feat to perform in aerial robots, especially without external positioning. However, it is routinely performed by biological fliers such as bees, flies, and bats. Our previous observations of landing behaviors in flies suggest an open-loop causal relationship between their putative visual cues and the kinematics of the aerial maneuvers executed. For example, the degree of rotational maneuver (the amount of body inversion prior to touchdown) and the amount of leg-assisted body swing both depend on the flies' initial body states while approaching the ceiling. In this work, inspired by the inverted landing behavior of flies, we used a physics-based simulation with experimental validation to systematically investigate how optimized inverted landing maneuvers depend on the initial approach velocities with varied magnitude and direction. This was done by analyzing the putative visual cues (that can be derived from onboard measurements) during optimal maneuvering trajectories. We identified a three-dimensional policy region, from which a mapping to a global inverted landing policy can be developed without the use of external positioning data. Through simulation, we also investigated the effects of an array of landing gear designs on the optimized landing performance and identified their advantages and disadvantages. The above results have been partially validated using limited experimental testing and will continue to inform and guide our future experiments, for example by applying the calculated global policy.

\end{abstract}

\section{INTRODUCTION}

Dynamic perching onto a ceiling surface is a feat many animal fliers (e.g. bats, flies  and bees \cite{bergou2015falling,liu2019flies,srinivasan2000honeybees,evangelista2010moment}) are  adept at; however, it is rarely achieved by robotic fliers in a self-reliant manner. Previous work has shown great promise for achieving dynamic perching on inclined surfaces, success of which, however, remained reliant on external motion tracking \cite{mellinger2012trajectory}. The ability of landing on inclined and inverted surfaces in robotic fliers will greatly expand their operational capacity, including sustained surveillance and inspection while also saving battery energy or even charging  \cite{mishra2020drone}\cite{kim2018drone}.

Similar to flying insects, execution of an inverted landing should use computationally efficient means within the capacity of embedded systems in nano-sized flying robots. Recent work by Mao and his coworkers has achieved landing on inclined surfaces without external positioning data \cite{mao2021aggressive}; however, their work is based on the computationally expensive process of generating and tracking optimal trajectories and does not allow for the fast reaction time or last-minute collision avoidance behaviors exhibited in insects \cite{liu2019flies}. By mimicking the sensory input of flies and focusing on their use of optical flow data, this fast reaction behavior can be better preserved and understood.

\begin{figure}[tpb]
    \centering
    \includegraphics[height=3.25in]{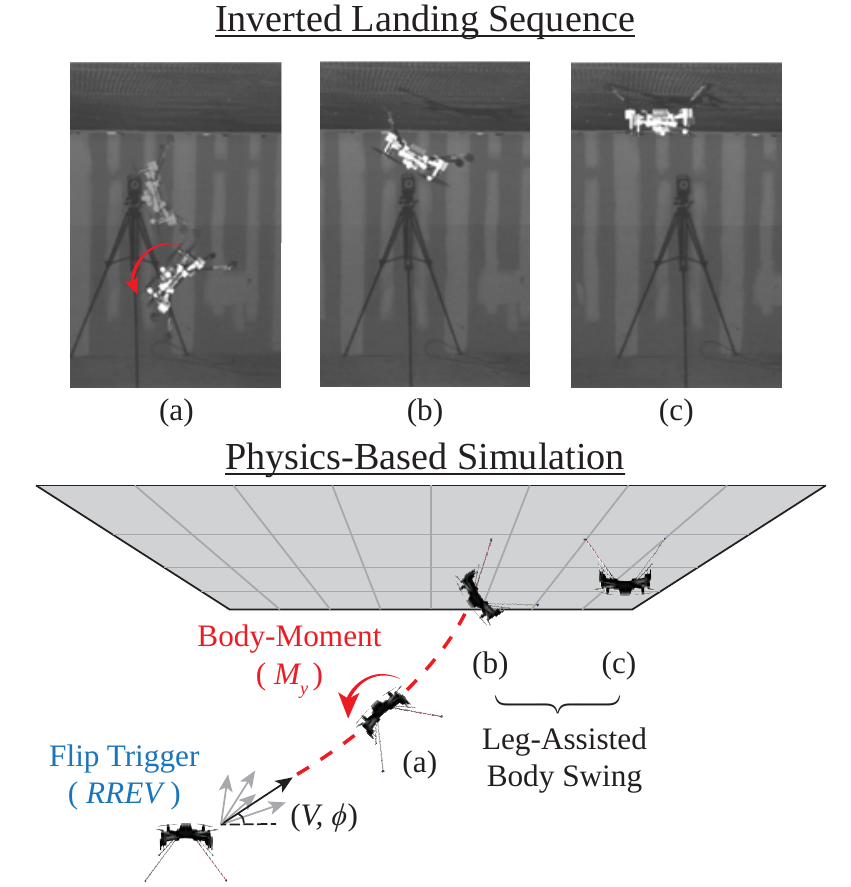}
    \caption{Inverted Landing Sequence: (a) Robot flies towards the ceiling at a constant velocity ($V$) and trajectory angle ($\phi$) until the rotational manuever is triggered. (b) Fore-legs make contact with the ceiling and remaining momentum is used to swing hind-legs up. (c) Final posture of inverted landing.}
    \label{fig:Experimental_Landing}
\end{figure}

In this study, to solve the problem of inverted landing with computational efficiency, we started by    identifying the minimal sensory cues from which a general control policy that maps sensory space to the appropriate action space can be designed. We first examined two predominant visual cues often used by insects: the Relative Retinal Expansion Velocity ($RREV$) of the approaching ceiling \cite{wagner1982flow}\cite{baird2013universal} and the fore/aft optical flow value ($OF_y$) \cite{srinivasan2000honeybees}. $RREV$ encodes a system's time to contact with the ceiling (assuming a constant approach velocity), while $OF_y$ encodes the ceiling's angular velocity (due to its relative fore/aft velocity) about the flying system.

It has been shown by Liu and his coworkers, that the combination of these variables lend themselves to the various inverted landing behaviors seen in flies \cite{liu2019flies}. A visual reference to these variables can be seen in Fig. \ref{fig:visual_cues} along with their mathematical expressions shown below, \cite{liu2019flies}

\begin{equation}
    \label{Eq:RREV}
    RREV = \frac{V_z}{d_{ceiling}},
\end{equation}
  
\begin{equation}
    \label{Eq:OFy}
    OF_y =   \frac{-V_x}{d_{ceiling}},
\end{equation}

where, $V_x$ and $V_z$ correspond to the horizontal and vertical velocity of the system respectively and $d_{ceiling}$ is the distance from the robot to the ceiling. In this work, the optical flow values were obtained based on these equations, in lieu of using a simulated or physical optical flow sensor, or an image processing algorithm  which will be used in follow-up work.

Another difficulty of achieving inverted landing in robotic fliers lies within their limited actuation capacity compared with their biological counterparts \cite{liu2019flies}\cite{cheng2016flight}. Biological fliers often exhibit superior force vectoring abilities that result in possibly fully or overly actuated flight. In addition, biological fliers can efficiently and quickly map visual and mechanosensory information to the motor signals that initiate, and control, a series of motion primitives via swift and large degree changes of aerodynamic forces \cite{liu2019flies} (Fig. \ref{fig:Experimental_Landing}). 

In this work, we implemented an EM-based Policy Hyper Parameter Exploration (EPHE) Reinforcement Learning algorithm \cite{wang2016based} to learn a series of optimized inverted landing control policies in a simulated environment; the method of which is also validated experimentally. The control policy generates the timing and magnitude of a rotational (or flip) maneuver over a large series of initial approaching velocities with varied magnitude and direction. By then analyzing all optimized policies and behaviors from each inverted landing at a particular approach velocity, a generalized control policy capable of generating near-optimal landings for an approaching velocity can be back-calculated. We also discuss the limitations of only using optical flow values for such a control policy and offer a way to resolve these problems with an augmented optical-flow space for future work. In addition, by repeating this process over various leg designs, we also assessed the effects of landing gear designs on the success of inverted landing. To verify the EPHE learning approach we completed a small set of tests under the vertical approach condition.

\begin{figure}[tpb]
    \centering
    \includegraphics[height=4cm,keepaspectratio]{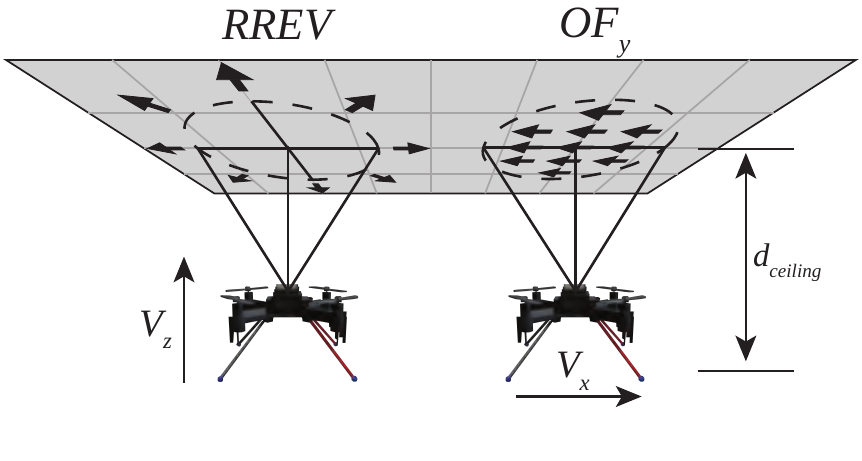}
    \caption{(a) Visual diagram representing the RREV. As the drone approaches the ceiling, observed points travel radially outwards at a corresponding radial velocity (b) Visual diagram representing optical flow about the $\bm{\mathrm{b_y}}$ axis. Feature points will travel across the field of view as the robot translates underneath.}
    \label{fig:visual_cues}
\end{figure}

The rest of the paper is organized as follows. Section II provides a description of our methodology and learning algorithms for optimizing landing policies. Section III validates our methods and results experimentally. Section IV details our results acquired via simulation and discusses our findings. Finally, Sec. V concludes the study and provides direction for our future work.

\section{METHODOLOGY}
From previous work on inverted landing in flies \cite{liu2020bio}, a series of motion primitives was executed for a successful inverted landing with hypothetically a desired angular pitch velocity ($\omega_d$) performed after the $RREV$ reached a triggering threshold ($RREV_{c}$) as it approaches the ceiling. However, we learned that directly emulating this strategy added complexity to the reinforcement learning framework with the addition of the quadrotor's flight controller and non-linear motor saturation limits to the environment. Often due to these saturation limits, and the small time scale of the maneuver, the desired angular velocities were not achievable and lead to inconclusive results. Instead, we circumvent the trajectory tracking by using a control policy to activate the flip with a desired body-moment ($M_{yd}$) about the quadrotor's $\bm{\mathrm{b_y}}$-axis; via adjusting the fore/aft motor thrusts.

For a given set of initial approach conditions, optimized control actions of the form ($RREV_{c}$, $M_{yd}$) could be found via simulation, which maximize the chances of a successful inverted landing. By determining these optimized policies over an array of approach conditions, a generalized policy valid across a large set of initial approach conditions could be formed. In each case of the simulation, flight control of the quadrotor was accomplished through a geometric tracking controller explained in detail through past works \cite{liu2020bio}\cite{lee2013nonlinear} with system properties determined by Forster et al. \cite{forster2015system}.

\subsection{Setup of Initial Approaching Velocities}

Each inverted landing was simulated for a constant approach condition (i.e., with a constant velocity magnitude $V$ and approach angle $\phi$), from which optimal control actions were learned (see Section C). Then, this process was iterated over a series of approach conditions, spanning  a parameter space with $V \in [0.25 - 4.0]$ m/s and $\phi \in [20^{\circ} - 90^{\circ}]$. 

Accordingly,  a collection of optimized control policy outputs were obtained for each $(V,\phi)$ combination. For each optimized inverted landing, the success percentage of four-legged landings, along with the optical flow and state values at the triggering time of the flip maneuver, were recorded from the final three converged episodes of learning (see Section C).

\subsection{Landing Gear Designs}

For each leg design configuration, the tips of the legs (or feet) were made to be adhesive to the ceiling and the hip joints modeled to be flexible with torsional spring and damping properties. When contact between the leg tips and the ceiling was made, the adhesive force was assumed to be strong enough to ensure a firm grip, with this behavior verified experimentally by using a VELCRO\textsuperscript{\texttrademark} connection between the landing gear feet and the ceiling surface. To determine the effect that various leg design parameters have on the inverted landing success rate and policy, a series of six designs were chosen which vary the leg length ($L$) and angle it makes with the $\mathbf{b}_z$ axis ($\psi$). An example of one such leg configuration can be seen in Fig. \ref{fig:body_coordinates_leg_design}b in which the leg length and angle are shown. The various leg dimensions that were tested can be seen in Table \ref{table:Leg_Dim}.

\begin{table}[h]
    \caption{Leg Design Configurations}
    
    \begin{center}
            \begin{tabular}{lccc}
                    \hline  Leg Design & $L$ (mm) & $\psi$ (deg) \\
                    \hline 
                    \text{ Extra Narrow-Short } & 50  & $0^{\circ}$ \\
                    \text{ Narrow-Short }       & 50  & $30^{\circ}$ \\
                    \text{ Wide-Short }         & 50  & $60^{\circ}$ \\
                    \text{ Extra Narrow-Long }  & 100  & $0^{\circ}$ \\
                    \text{ Narrow-Long }        & 100  & $30^{\circ}$ \\
                    \text{ Wide-Long }          & 100  & $60^{\circ}$ \\
                    \hline
            \end{tabular}
    \end{center}
    \label{table:Leg_Dim}
\end{table}

\subsection{Optimization of Landing Policy with EPHE Algorithm}

For each initial approach velocity, we aim to directly optimize the control actions, in the terms ($RREV_{c}, M_{yd}$), which maximize landing success instead of a general control policy \textit{per se}; which will be back-calculated based on all tested initial conditions. The $RREV_{c}$ corresponds with the timing of the flip and $M_{yd}$ to a desired body-moment about the robot's center of mass (COM).

Optimization of these parameters was performed via the EPHE algorithm which exhibited quick convergence due to its adaptive learning rate. The Gaussian distributions used in the EPHE algorithm were of the form, $\bm{O} = \left[O_{RREV}, O_{M_{yd}}\right]$ where $\bm{O} \sim  \mathcal{N} (\bm{\mu}, I\bm{\sigma^2})$ with $\bm{\mu} = \left[\mu_{RREV}, \mu_{My} \right]$ and $\bm{\sigma} = \left[\sigma_{RREV}, \sigma_{My}\right]$. For details of the EPHE algorithm, refer to the work done by Wang et. al \cite{wang2016based}.

\begin{figure}[tp]
    \centering
    \includegraphics[width=0.85\linewidth]{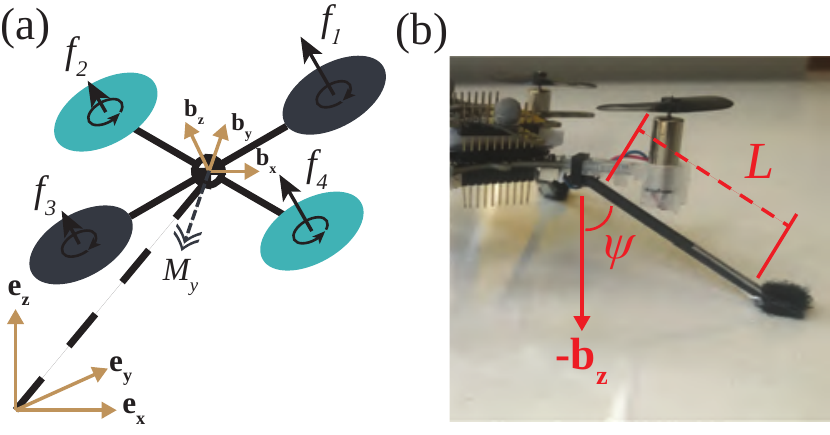}
    \caption{(a) Quadrotor coordinate reference frame. Flip manuevers are activated through a negative rotational moment generated about the $\bm{\mathrm{b_y}}$ axis via the increased thrust in motors 1 and 4. (b) Diagram showing the parameters for leg angle ($\psi$) and leg length ($L$).}
    \label{fig:body_coordinates_leg_design}
\end{figure}

For each rollout, the quadrotor started in an initial hovering state of $\bm{x_0} = \left[0,0,d_0\right]$, 
$\bm{v_0}=\left[0,0,0\right]$, 
$\bm{R_0}=\left[\bm{e_x};\bm{e_y};\bm{e_z}\right]$, 
$\bm{\Omega_0}=\left[0,0,0\right]$, where $\bm{x}$,$\bm{v}$,$\bm{R}$, and $\bm{\Omega}$ are the initial position, velocity, orientation, and angular rates respectively. The term, $d_0$, is the initial hovering distance of the quadrotor from the ceiling surface, and the vectors $\bm{e_x}$, $\bm{e_y}$, $\bm{e_z}$ represent the global coordinate axes (Fig. \ref{fig:body_coordinates_leg_design}).

A successful inverted landing was characterized by several conditions which were used to design each rollout's reward: the minimum achieved distance of the quadrotor from the ceiling ($d_{ceil}$), as in (\ref{Eq:r_height}); a dense reward calculated from the quadrotor's orientation about its $\bm{\mathrm{b_y}}$-axis at impact ($\theta_{impact}$), which provides a smooth gradient encouraging the robot to finish the rollout in an inverted state, represented as (\ref{Eq:r_theta}); a sparse reward term that increases with the number of legs contacting the ceiling, presented  in (\ref{Eq:r_legs}). These terms were summed together in the form $R = r_{d_{ceil}}+r_{\theta} + r_{legs}$ with the additional penalty of $r_{legs} = r_{legs}/2$ when the body or rotors make contact with the ceiling surface. 

\begin{equation}
    \label{Eq:r_height}
    r_{d_{ceil}} = \frac{1}{{min(d_{ceil})}},
\end{equation}
\begin{equation}
    \label{Eq:r_theta}
    r_{\theta} = 
            \begin{cases} 
            10 \cdot \frac{|\theta_{impact}|}{90^{\circ}} & 0^{\circ} \le |\theta_{impact}| < 90^{\circ} \\
            10 & 90^{\circ} \le |\theta_{impact}| \le 180^{\circ} 
           
            \end{cases},
\end{equation}
\begin{equation}
    \label{Eq:r_legs}
    r_{legs} = 
            \begin{cases} 
            150 & N_{legs} =3 \ || \ 4 \\
            50 & N_{legs} = 1 \ || \ 2 \\
            0 & N_{legs} = 0 \\
            
            \end{cases},
\end{equation}

To begin each rollout, a deterministic policy of the form $\bm{\theta} = \left[\theta_{RREV},\theta_{M_{yd}}\right]$ was sampled from the distribution $\bm{O}$ and used to execute the landing maneuver. The quadrotor then followed a constant velocity trajectory at the desired speed $V$ and flight angle $\phi$, thereby approaching the ceiling until the controller receives an $RREV$ value greater than $\theta_{RREV}$. At which point, the controller executed the desired body rotational moment $\theta_{My}$ about the quadrotor's $\bm{\mathrm{b_y}}$-axis by increasing the thrust of the front rotors (Fig \ref{fig:body_coordinates_leg_design}). The rollout would then end after a timeout threshold was exceeded or the landing was successful, and its reward was then calculated.

In each learning trial, convergence on the distribution $\bm{O}$ was achieved with the reward being maximized by the EPHE algorithm and training being done by selecting the best K returns in each episode of N rollouts. Upon convergence, a near deterministic set of parameters was obtained which maximized the probability of a successful landing for a given initial velocity. In each case, convergence to a deterministic value would typically occur within 100 rollouts. This process was then repeated five times for each set of flight conditions ($V$,$\phi$) inside of $V \in [0.25 - 4.0]$ m/s and $\phi \in [20^{\circ} - 90^{\circ}]$ to create a sizeable parameter space of successful policies.

\section{RESULTS \& DISCUSSION}

\subsection{Validation of Simulation - Experimental learning of Inverted Landing with Vertical Approaching Velocity}

The inverted landing experiments were performed on a nano-sized quadcopter (Bitcraze Crazyflie 2.1) equipped with four upgraded brushed DC motors. This small flying robot was chosen because of its low mass, high maneuverability, and open-source firmware which lends well to experimental testing. Communication to the drone was done through the Crazyswarm package \cite{preiss2017crazyswarm} which served as a bridge between Robot Operating System (ROS) communication messages and the Crazyflie Real Time Protocol (CRTP) operating inside the system firmware. To ensure good adhesion each 3D printed leg was affixed with a VELCRO\textsuperscript{\texttrademark} pad on its ventral side of the foot which could join to the ceiling panel under very light but sufficient contact.

Real time position data was streamed to the Crazyflie drone using a Vicon motion capture system operating at 100 Hz. This data was used exclusively for experiment monitoring purposes, such as ensuring the proper velocity was maintained. As well as, obtaining the ground truth of the optical flow values $RREV$ and $OF_y$, expressed in \eqref{Eq:RREV} and \eqref{Eq:OFy} respectively. Which serve as a proxy of those recorded with a mounted camera or optical flow sensor. Throughout our experimental results, any external data used was for the purpose of data recording and emulating onboard sensory measurements.

To experimentally validate the learning process in simulated landings, experimental landing tests were completed in identical fashion but limited to a vertical approach velocity and configured with the Wide-Short leg design. For each rollout, the quadrotor started in a hover state at a distance of 1.7m from the ceiling and it then followed a pre-programmed trajectory that accelerated the robot upwards to a constant vertical velocity of 2.50 m/s towards the ceiling. Through implementing the EPHE algorithm as described previously, by which it sampled $RREV_{c}$ and $M_{yd}$ values from their respective Gaussian distributions and rewarded each rollout accordingly, an experimentally validated inverted landing strategy can be achieved with a convergence towards an optimized policy. Examples of experimental inverted landings and the policy convergence process (via simulation) can be found at ({\fontfamily{pcr}\selectfont https://youtu.be/H3dNZyrKxd0}).

\begin{figure}[tpb]
    \centering
    \includegraphics[height=4cm,keepaspectratio]{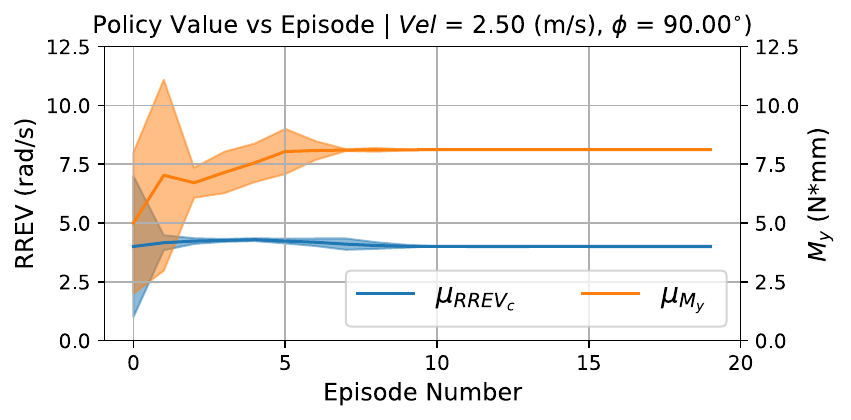}
    \caption{Policy convergence plot showing convergence on a successful inverted landing policy in an experimental trial. The plot shows the convergence of the learned policy parameters $RREV_{c}$ and $M_{yd}$. Where each shaded region represents the $\mu_i\pm 2\sigma_i$ distribution that rollouts in each episode are sampled from.}
    \label{fig:RL_Experiment}
\end{figure}

Our initial policy distributions started with a $\bm{\mu} = \left[4.0,5.0 \right]$
and a $\bm{\sigma} = \left[1.50,1.50\right]$, which were chosen based on the valid solutions provided by the simulation. Due to the nature of random Gaussian sampling provided in the learning process, combinations of values in the $2\sigma_i$ range around each $\mu_i$ were explored and tested, which assists in removing any potential biases from the initial policy distributions. The learning process successfully converged on an optimized policy which resulted in a landing strategy with a 93.75\%  success rate for achieving a four-leg/fully inverted landing. The optimized policy converged on an $RREV_{c} = 4.0$ rad/s and $M_{yd}=8.12$ N*mm which was close to the results predicted by the simulation; in which the converged spread of polices were often around $RREV_{c} = 4.64$ rad/s and $M_{yd}=8.09$ N*mm. Plots showing the parameter convergence of the experimental trial can be seen in Fig. \ref{fig:RL_Experiment}. The above results validated that our design of control policy and the learning process can be reliably translated from simulation to physical experiments.

\subsection{Simulation Results}

\begin{figure}[tp]
    \centering
    \includegraphics[width=0.75\columnwidth]{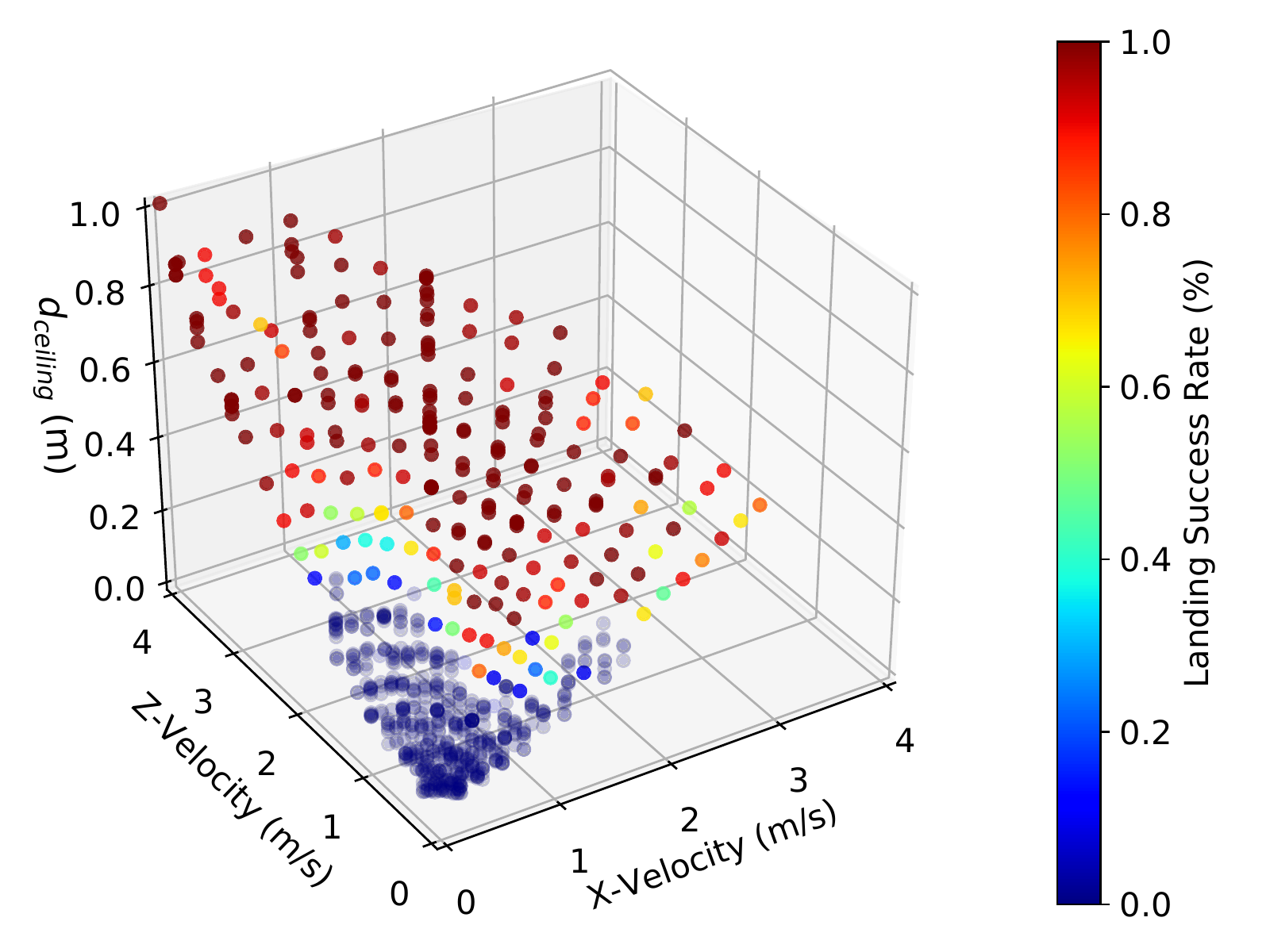}
    \caption{State-space of the simulated flight conditions for the Wide-Short leg configuration. Each data-point signifies the tested velocities and final $d_{ceiling}$ at which the optimized flip manuever was initiated. A clear region can be seen which separates the successful landings from the failed landings.}
    \label{fig:State_Space_Flip_Region}
\end{figure}

One of the objectives of our simulation tests is to obtain a generalized landing policy that efficiently maps the available onboard measurement data to a resulting flip action; therefore simulation data including the directly optimized policy actions from optimized inverted landing over a large array of initial approaching velocities was analyzed. As shown in Fig. \ref{fig:State_Space_Flip_Region}, there exists a clear policy region inside of the quadrotor's state-space, which corresponds to optimized inverted landings with high success rates. Each point shown is a tested set of ($V, \phi$) initial approach conditions mapped to a Cartesian pair ($V_x,V_z$), along with the corresponding distance from the ceiling ($d_{ceiling}$) at which the flip maneuver was executed; the color represents the landing success rate. Note that there is a clear separation between the failed and successful landings, the region of the latter is well enclosed and relatively smooth, therefore can be  used to extract a general and compact landing policy. 

\begin{figure}[tp]
    \centering
    \includegraphics[width=0.70\columnwidth]{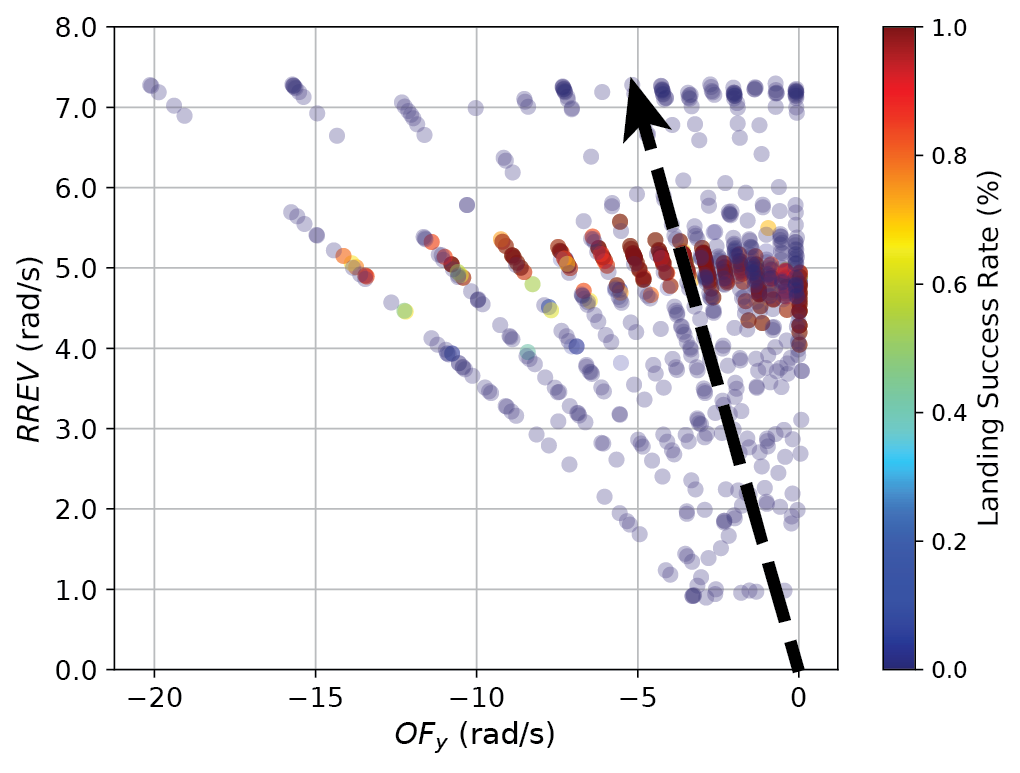}
    \caption{Optical Flow-Space of the simulated flight conditions for the Wide-Short leg configuration. Each point here identifies the ($RREV$, $OF_y$) pair when an optimized flip manuever was triggered. Failed landings pervade the entire optical flow-space with no clear relation that can separate the high landing rate combinations from the low landing rate combinations overlapping them. The black line signifies an example flight trajectory.}
    \label{fig:OF_Space_Flip_Region}
\end{figure}

However, we next plotted the policy region in the robot's emulated optical flow-space; which consists primarily of $RREV$ and $OF_y$ (Fig. \ref{fig:OF_Space_Flip_Region}). Here, no enclosed region or correlation exists that can separate the flip triggering points with a high success rate from those with low success rate. Therefore, it is unsuccessful to locate a general control policy exclusively based on the two optical flow variables as the robot translates through its  optical flow-space.

This failure indicates that the optimal inverted landing or control policy depends critically on the distance estimation or distance-related information, which is however redundant or unresolved with merely $RREV$ and $OF_y$  optical flow variables, see (\ref{Eq:RREV}) and (\ref{Eq:OFy}). 
For example, a high approach velocity and a large distance from the ceiling can give the same combination of ($RREV$, $OF_y$) as a low approach velocity and a small distance from the ceiling; despite them both requiring vastly different landing strategies. Therefore, an additional state or perception term must either be sensed directly or estimated from onboard measurements to form a generalized landing policy. 

\begin{figure}[tp]
    \centering
    \includegraphics[width=0.85\columnwidth]{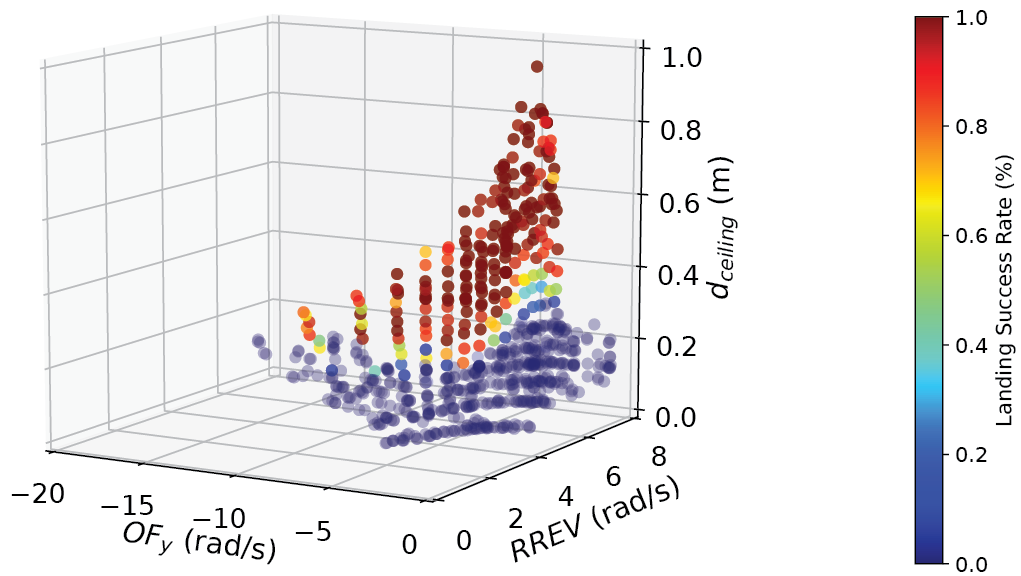}

    \caption{A clear region separating high and low landing rates can be identified when plotting the optimized flip conditions from Fig. 6 with augmented information about $d_{ceiling}$.}

    \label{fig:Wide-Short_Policy_Volume_LR}
\end{figure}

\begin{figure}[tp]
    \centering

    \includegraphics[width=0.85\columnwidth]{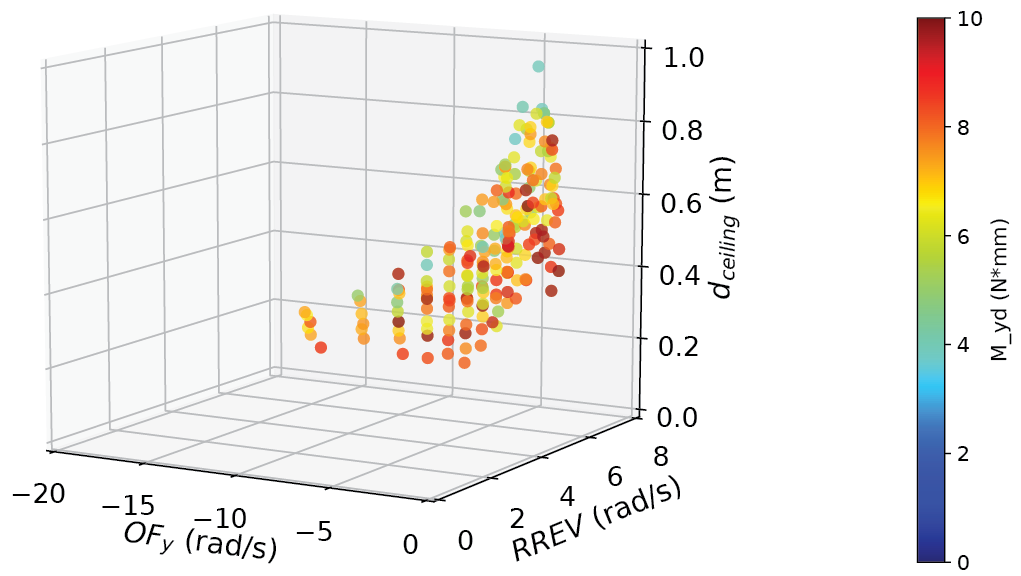}

    \caption{A non-linear policy mapping for $M_{yd}$ can be extracted from the same measurement space as Fig. \ref{fig:Wide-Short_Policy_Volume_LR} by looking at only the successful landings.}

    \label{fig:Wide-Short_Policy_Volume_My_d}
\end{figure}

To resolve this problem, one can augment the optical flow space by additional sensory or derived-perceptual variable(s), and the simplistic solutions seem to estimate the distance or linear velocity and add it to the control policy. It can be seen in Fig. \ref{fig:Wide-Short_Policy_Volume_LR} that by including this $d_{ceiling}$ term a clear separation between landing success rates can be re-established and there exists an enclosed policy region. In addition, Fig. \ref{fig:Wide-Short_Policy_Volume_My_d} shows that the mapping from the policy region to control action is highly nonlinear. Therefore, one may either need to further augment the sensory space to obtain a more linear mapping, or using a nonlinear function approximator (such as neural network) to model this mapping.

\subsection{Method of Distance Estimation}

\begin{figure*}
    \centering
    \includegraphics[width=0.95\textwidth]{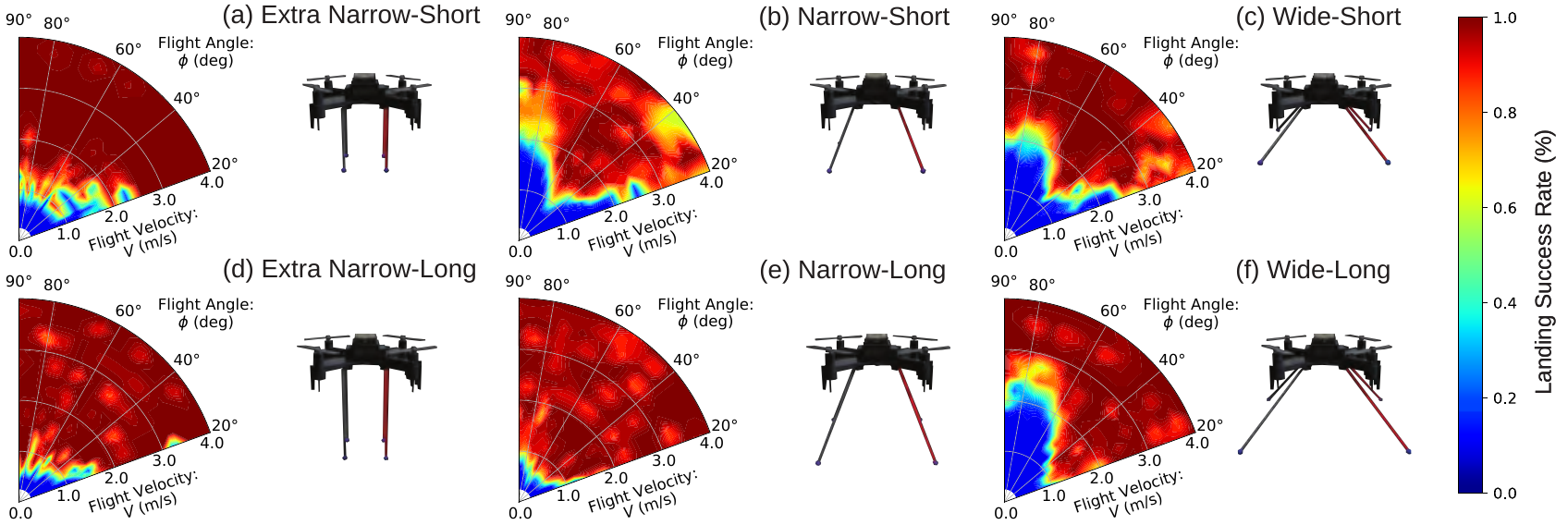}
    \caption{Landing success rate plots for each of the simulated leg configurations over the tested parameter space $V \in [0.25 - 4.0]$ m/s and $\phi \in [20^{\circ} - 90^{\circ}]$.}
    \label{fig:Polar_Plots}
\end{figure*}

Therefore, to develop a general compact policy implementable with onboard resources, one solution is to estimate the distance to the ceiling and augment the optical-flow space. As presented by Breugel et al. \cite{van2014monocular}, a feasible method of distance estimation based on optical flow values is to also include the robot's vertical acceleration from an onboard accelerometer. By incorporating acceleration with the time derivative of the $RREV$, an expression to estimate $d_{ceiling}$ can be derived (\ref{Eq:d_ceiling_dRREV})

\begin{equation}
    \label{Eq:d_ceiling_dRREV}
    d_{ceiling} = \frac{\ddot{z}}{\dot{RREV}-RREV^2}.
\end{equation}

However, as the acceleration approaches zero (i.e., with an constant approach velocity), this expression quickly becomes undefined, thus requiring the robot to have an active acceleration towards the ceiling during the approach. Remarkably, this is consistent with past observations that biological fliers will often approach a landing surface through maintaining a constant ${RREV}$ \cite{evangelista2010moment}\cite{baird2013universal}\cite{van2012visual} which requires a non-zero vertical acceleration, thus allowing this distance measurement to be reliably estimated. In addition, our past work on inverted landing in flies also revealed consistent acceleration towards the ceiling before the flip (rotational) maneuvers were triggered \cite{liu2019flies}.

By then analyzing a landing trajectory in this augmented optical flow space, the segments that intersect or are sufficiently close to the policy region can be used to calculate a desired policy action (i.e., triggering timing and maneuver moment) that results in a successful inverted landing. The policy region identified here can also be used in future work for trajectory planning to have the robot enter the region via a feasible path.

Also note that there exists some degree of mechanical robustness from the leg designs, consequently, there does not exist a one-to-one deterministic relation between the sensor (perceptual) information in the policy region and the requisite $M_y$ value. Therefore, the policy can be justifiably probabilistic, as a distribution of $M_y$ values exist that allow for a successful inverted landing with a viable range of body-angles at impact which are capable of utilizing the drone's momentum to swing the remaining legs up to the ceiling surface. The specific policies and landing behaviors may vary greatly among leg designs.

\subsection{Simulation Results - Effects of Landing Gear Designs}

Figure \ref{fig:Polar_Plots} summarizes the success rates of optimized inverted landing for six different landing gear designs, each with varied approach velocity magnitudes and direction. It can be seen that high approach velocity is often most advantageous for higher success rates, while the landing gear design has relatively minor effects in bias towards certain approach directions. For example, the Wide-Long leg design (Fig. \ref{fig:Polar_Plots}f) favors shallower approach angles where $\phi$ is between $20^{\circ}$ and $50^{\circ}$, while the Narrow-Short design favors an approach angle near $\phi= 60^{\circ}$. Overall, the Long leg designs largely outperformed the Short ones at low flight speeds. Nonetheless, all designs had largely consistent performance and high success rate at high approach velocities, once the policy was optimized. 

Across the various leg designs, a consistent optimal strategy was observed, where-by leveraging a proper body impact angle the body momentum about the impact point can be utilized to swing the remaining legs up and make full contact with the ceiling. Therefore, a higher linear momentum or optimized impact angle would have the dominant effect on determining a successful inverted landing.

Results also show that landing gears with long and narrow legs largely outperform other designs at low approach velocities. This is primarily due to the large displacement of the center of mass (COM) from the impact point causing a larger angular moment and pendulum-like swing, while also requiring a smaller degree of rotation to make the rear legs contact the ceiling. However, the primary disadvantages of the Narrow-Long leg designs are the increased amount of body rotation needed to have a sufficient body-angle at impact and the larger moment of inertia due to the longer legs. Both aspects will require more rotation time and possibly limit the viable impact conditions, these aspects need to be investigated more comprehensively in experimental settings.

The pendulum-like body swing also became more prominent in the Wide-Long/Wide-Short designs undergoing a shallow approach angle or higher forward velocity. In these scenarios, as the quadrotor had a lower degree of body rotation at impact, the linear momentum could be more efficiently transferred into the COM near the bottom of its pendulum arc and swing up the rear legs to make contact with the ceiling.

\section{CONCLUSION AND FUTURE WORK}

In this work, we used a physics-based simulation and reinforcement learning algorithm to acquire optimized inverted landing actions employed in a quadrotor robot under a large array of initial approach conditions. By analyzing the system's states and actions at the time of triggering the rotational (flip) maneuver, and validating our methods experimentally, a general optical flow policy in augmented optical flow space emerged from emulated sensory information. Continued work will benefit from the corresponding control policy region using compact function approximation and mapping it to control actions based on real-time onboard sensors, thereby allowing self-sufficient inverted landings over a large multitude of trajectories passing through this policy space.

This framework opens opportunities to future path-planning and trajectory optimization problems targeted at reaching the identified control policy region that can result in truly optimized landings which minimize impact force, energy usage, or maximize overall landing robustness. Further, the identified control policy region for successful landing can give clear information as to when a collision avoidance maneuver is needed due to insufficient policy conditions, which can be used to prevent damage from a failed landing. 

Results about the performance of various landing gear designs overwhelmingly suggested that the flight velocity is the greatest contributing factor to landing success rate, while acceleration towards the ceiling also assists the distance estimation that better informs the control policy. Landing gears have relatively  minor effects on the landing success once the control policy is optimized; however, their effects on the landing policy itself needs further investigation.

\addtolength{\textheight}{-12cm}   




\section*{ACKNOWLEDGMENT}

This research was supported by the National Science
Foundation (IIS-1815519 and CMMI-1554429).


\bibliography{refs.bib}

\bibliographystyle{IEEEtran}

\end{document}